\documentclass[11]{amsart}
\usepackage{graphicx}
\usepackage{textcomp}
\usepackage{xcolor}
\usepackage{cite}
\usepackage[foot]{amsaddr}
\usepackage[utf8]{inputenc}
\usepackage{graphicx}
\usepackage{xcolor}

\newcommand\skipthis[1]{}

\title{Baseline Computation for Attribution Methods Based on Interpolated Inputs}

\author[1]{Miguel Lerma}
\address[1]{Northwestern University, Evanston, USA}
\email[1]{mlerma@math.northwestern.edu}
\author[2]{Mirtha Lucas}
\address[2]{DePaul University, Chicago, USA}
\email[2]{mlucas3@depaul.edu}

\date{\today}

\begin{document}

\begin{abstract}
We discuss a way to find a well behaved baseline for attribution methods
that work by feeding a neural network with a sequence of interpolated
inputs between two given inputs. Then, we test it with our novel Riemann-Stieltjes
Integrated Gradient-weighted Class Activation Mapping (RSI-Grad-CAM) attribution method.
\end{abstract}

\maketitle

\section{Introduction}

Providing explanations to the output of deep neural networks
is a central problem in Explainable Artificial Intelligence (XAI).
For this purpose a wide variety of methods have been proposed
\cite{kokhlikyan2020captum},
but here we will focus on gradient based methods
such as the Gradient-weighted Class Activation Mapping
(Grad-CAM)  \cite{selvaraju2017grad}, and its variations
(see e.g. \cite{chattopadhyay2018gradplus}). 
In general, gradient based methods can be applied to classifier
networks such as VGG19 \cite{bengio2012deep, russakovsky2015, simonayan2015}
and ResNet50 \cite{he2015resnet}. 
They work by using backpropagation of gradients to determine in what 
extent the output of the network depends on the activation of its units.
More specifically, if $A^k_{ij}$ represents the activation of a unit 
placed in position $i,j$ of feature map $k$
of a given layer, and $y^c$ is the output of the network corresponding to
class $c$, then the partial derivative $\frac{\partial y^c}{\partial A^k_{ij}}$
can be interpreted as how much the output of the network depends
on the activation of the given unit.  Gradient based methods use 
the value of those partial derivatives (gradients) to build \emph{heatmaps}
that can be mapped to the input of the network to determine what areas of the
input are responsible for the network output. Also, the computation of the
gradients can be performed with the same tools used to train the network,
i.e., the backpropagation algorithm.

In spite of its popularity, gradient based methods have a drawback, namely
the vanishing gradients problem that arises when some units work at or close to
their saturation, their outputs do not change any more and the gradients
become zero or near zero. We will examine this problem in more detail
in section \ref{S2}, followed by the description of two 
attribution methods intended to work around the problem, namely
Integrated Gradients (IG) \cite{sattarzadeh2021igcam}, and our novel
Riemann-Stieltjes Integrated Gradient-weighted Class Activation Mapping
(RSI-Grad-CAM).\footnote{Submitted for publication.}

\section{Attribution methods based on interpolated inputs}
\label{S2}

\begin{figure}[htb]
\centering
\includegraphics[width=3in]{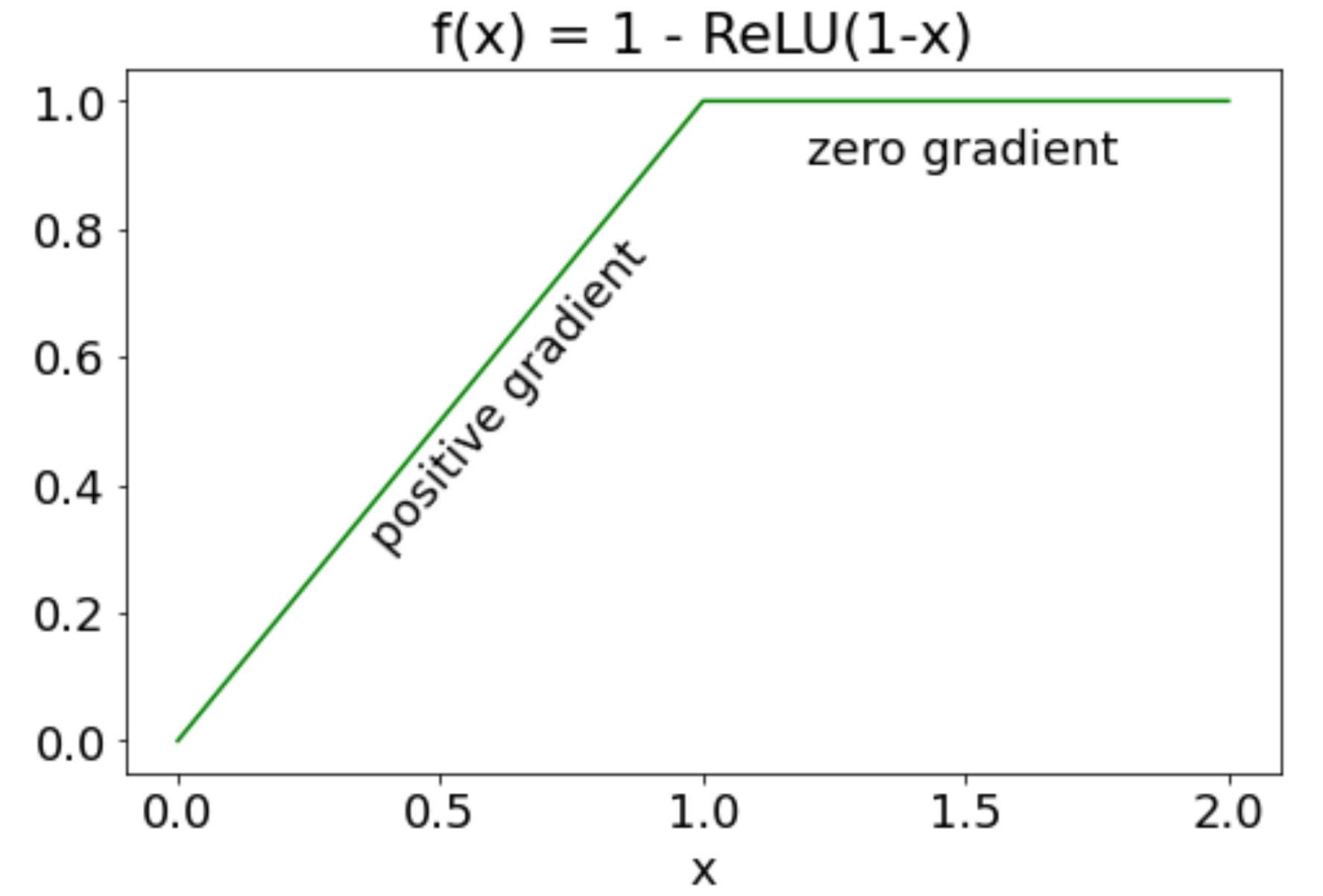}
\caption{Example of function with the vanishing gradient problem.}\label{f:one_relu_net}
\end{figure}

As mentioned in the introduction, gradient based methods are affected by
the vanishing gradients problem. This can be illustrated with a unit with
input $x$ and output $f(x) = 1 - \text{ReLU}(1-x)$, where 
$\text{ReLU}(x) = \max(x,0)$---see figure (\ref{f:one_relu_net}).
For such unit the output would steadily increase as $x$ varies from $0$ to $1$
and then remain constant for $x>1$, hence $f'(x) = 0$.  A gradient based method
applied to this unit would attribute a value of zero to the contribution 
of its output to its output when $x>1$, which clashes with the fact that 
$f(x)$ does in fact increase by $1$ when $x$ varies from $0$ to $2$.
In order to remediate this problem we can replace the derivative of $f$ with 
the integral of the derivative to obtain the actual change experienced by the function. 
This yields:
\begin{equation}
  I = \int_0^2 f'(x) \, dx = 1 \,.
\end{equation}
Note that the result depends on the starting and ending points, namely
$0$ and $2$. While the latest will be seen as a given (attribution of input to output at $x=2$),
the former (the starting point of the integration---the "baseline")
is somehow arbitrary and may need to be 
justified or picked carefully.  We will come back to this issue later.

\subsection{Integrated Gradients}

Here we describe a first method of attribution designed to avoid the problem of
vanishing gradients, namely Integrated Gradients, introduced in \cite{sundararajan2017ig}.

Assume we have a network intended to classify images.
The main idea of this method is as follows:
instead of inputting a single image we use a
sequence of interpolated images between a baseline $x'$ and the given
image $x$. Each interpolated image is a combination
$x' + \alpha (x-x')$ with $0 \leq \alpha \leq 1$.  Then, the gradient
of the network output with respect to each input pixel $x_i$
is integrated as shown in equation (\ref{e:igint}).

\begin{equation}\label{e:igint}
  \texttt{IntegratedGrads}_i(x) ::= 
  (x_i - x'_i) \times \int_{\alpha=0}^{\alpha=1}
  \frac{\partial F(x' + \alpha \times (x - x'))}{\partial x_i} \, d\alpha \,.
\end{equation}

In the equation, the factor 
$(x-x')$ appears when using as variables of
integration the pixel values of the interpolated image, so that
the differential within the integral is
$d(x' + \alpha \times (x - x')) = (x-x') \times d\alpha$.
Since it does not depend on $\alpha$, the factor $(x-x')$ can be taken 
outside the integral.

In practice the integral can be approximated numerically with a summation:
\begin{equation}\label{e:igapprox}
 \texttt{IntegratedGrads}_i^{approx}(x) ::= 
 (x_i - x'_i) \sum_{\ell=1}^m
 \frac{F(x' + \frac{\ell}{m}\times (x-x'))}{\partial x_i} \times \frac{1}{m} \,,
\end{equation}
where $m$ represents the number of interpolation steps. This parameter
$m$ can be adjusted by experimentation, although it is recommended to give it
a value between 50 and 200.

A problem with the Integrated Gradients method is that it is designed
for working with the network inputs and may miss features captured
at hidden layers.  We next describe a way to fix this problem.

\subsection{RSI-Grad-CAM}
Our RSI-Grad-CAM is an integrated gradients method that can be applied to
any layer rather than just the network input.
It still requires feeding the network with a sequence of interpolated
images, but the gradients are integrated at a hidden layer rather than the input.

More specifically, first we must pick a convolutional layer
$A$, which is composed of a number of feature maps $A^1, A^2, \dots, A^{N}$,
all of them with the same dimensions.  
If $A^k$ is the $k$-th feature map of the picked
layer, and  $A_{ij}^k$ is the activation of the unit in the
position $(i,j)$ of the $k$-th feature map,
then, a localization map or ``heatmap'' can be obtained
by combining the feature maps of the chosen layer
using weights $w_k^c$ that capture the contribution of the $k$-th
feature map to the output $y^c$ of the network corresponding to class~$c$.
There are various ways to compute the weights $w_k^c$.
For example Grad-CAM, introduced in \cite{selvaraju2017grad}, uses the
gradient of the selected output $y^c$ with respect to the activations
$A_{ij}^k$ averaged over each feature map, as shown in equation (\ref{e:gradcam_weights}). Here $Z$ is the size (number of
units) of the feature map.

\begin{equation}\label{e:gradcam_weights}
  w_k^c = \overbrace{\frac{1}{Z} \sum_{i}\sum_{j}}^{\text{global average pooling}}
  \hskip -30pt
  \underbrace{\frac{\partial y^c}{\partial A_{ij}^k}}_{\text{gradients via backprop}}
  \,.
\end{equation}

On the other hand, our RSI-Grad-CAM computes the weights 
$w_k^c$ using integrated gradients in the following way.
First we need to pick a baseline input $I_0$ (when working 
with images $I_0$ is typically a black image).
Then, given an input $I$, we consider the 
path given in parametric form $I(\alpha) = I_0 + \alpha (I - I_0)$,
where $\alpha$ varies between $0$ and $1$,
so that $I(0) =I_0$ (baseline) and $I(1) = I$ (the given input). 
When feeding the network with input $I(\alpha)$, the output 
corresponding to class $c$ will be $y^c(\alpha)$, and the activations of
the feature map $k$ of layer $A$ will be $A_{ij}^k(\alpha)$.
Then, we compute the weights by averaging the integral of 
gradients over the feature map,
as shown in equation (\ref{e:igcam_sint}).

\begin{equation}\label{e:igcam_sint}
  w_k^c = \frac{1}{Z}
  \sum_{i,j} \int_{\alpha=0}^{\alpha=1} \frac{\partial y^c(\alpha)}{\partial A_{ij}^k} \, d A_{ij}^k(\alpha)
  \,.
\end{equation}

The integral occurring in equation (\ref{e:igcam_sint})
is the Riemann-Stieltjes integral of function 
$\partial y^c(\alpha)/\partial A_{ij}^k$
with respect to function $A_{ij}^k(\alpha)$ (see \cite{protter1991integral}).
For computational purposes this integral can be approximated with 
a Riemann-Stieltjes sum:

\begin{equation}\label{e:igcam_ssum}
  w_k^c = \frac{1}{Z} \sum_{i,j} \left(\sum_{\ell=1}^m
    \Biggl\{\frac{\partial y^c(\alpha_{\ell})}{\partial A_{ij}^k} 
    \times \Delta A(\alpha_{\ell}) \Biggr\}\right)
    \,.
\end{equation}
where 
$\Delta A(\alpha_{\ell}) = A_{ij}^k(\alpha_{\ell}) - A_{ij}^k(\alpha_{\ell-1})$,
$\alpha_{\ell} = \ell/m$, and $m$ is the number of interpolation steps.

The next step consists of combining the feature maps $A^k$ with the
weights computed above, as shown in equation (\ref{e:heatmap}).  Note
that the combination is also followed by a Rectified Linear function
$\text{ReLU}(x) = \text{max}(x,0)$,
because we are interested only in the features that have a positive
influence on the class of interest.

\begin{equation}\label{e:heatmap}
  L^c =
  \text{ReLU} \underbrace{\Biggl(\sum_k w_k^c A^k\Biggr)}_{\text{linear combination}}
  \,.
\end{equation}

After the heatmap has been produced, it can be normalized and
upsampled via bilinear interpolation to the size of the original image,
and overlapped with it to highlight the areas of the input image that
contribute to the network output corresponding to the chosen class.

\section{Baseline Computation}

The two methods discussed require the choice of a baseline input $I_0$
as the starting point for the image interpolation.  It is natural to think that
the absence of image may  be given by a black image, but this may not 
work well in all cases---see e.g. \cite{sturmfels2020} for a detailed discussion.
The problem is hard and far from solved, but we can at least require a natural
condition on the baseline: when fed to the network the output should be
a uniform vector of probabilities, i.e, all classes should be assigned the same
probability.  The heuristics behind this condition is that the baseline
should not be representative of any class, and the output should be equivalent
to an ``I don't know'' to the question ``to which class does this image belong?''
This often is not the case, e.g. using a VGG19 network pre-retrained with ImageNet
(1000 classes) one would expect that ``I don't know" would be equivalent to
outputting a probability of $1/1000 = 0.001$ for each class, but that is not what
happens when the network is fed with a completely black image. 
The actual input of the network is a probability vector with its elements ranging from
$1 \times 10^{-5}$ to $0.76$, with the maximum probabilities assigned to 
'matchstick' (0.076), 'nematode' (0.057), and 'digital clock' (0.03).

One way to fix this anomaly is to perturb the chosen baseline so to force the desired 
uniform probability distribution vector as the output of the network. The same procedure that
we are going to describe here can be used if we have good reasons to believe that the 
class probability distribution should follow some given non uniform
distribution, just replace the target uniform probability vector with
whatever vector represents the desired distribution.

So, if we denote  $\mathbf{o}_{b}=$ the output of the network when fed with the baseline,
and $\mathbf{o}_{t}=$ the desired target output, then the problem can be posed as a
minimization problem with some loss function $\mathcal{L}_{out}(\mathbf{o}_{b},\mathbf{o}_{t})$
that attains its minimum when $\mathbf{o}_{b} = \mathbf{o}_{t}$.
At the same time we want to avoid the baseline to change too much and become a tensor too far away from the
image space of interest, so we must add a second loss function $\mathcal{L}_b(\mathbf{b},\mathbf{b}_0)$,
where $\mathbf{b}=$ baseline, $\mathbf{b}_0=$ original baseline (e.g. a black image).
The final loss function is:
\begin{equation}\label{e:loss}
    \mathcal{L} = \lambda \mathcal{L}_{out} + (1 - \lambda) \mathcal{L}_0
    \,.
\end{equation}
where $\lambda$ is a parameter between $0$ and $1$ used to assign more or less weight to each term.

The problem can now be posed as training a network with one input unit set to $1.0$, and a fully connected
layer L  with its initial weights equal to the pixel values of the initial baseline $\mathbf{b}_0$. The output of
this layer is fed to the network N. Next, we train the composite network N+L using the loss function 
in equation (\ref{e:loss}).

\subsection{Results}
In our experiments we used $\mathcal{L}_0=$ mean square error, and $\mathcal{L}_{out}=$ mean square error,
categorical crossentropy, and a custom loss equal to $|\mathbf{o}_{b} - \mathbf{o}_{t}|_{max}$,
where $|(x_1,x_2,\dots,x_n)|_{max} = \max(x_1,x_2,\dots,x_n)$. The best results were obtained for
$\mathcal{L}_{out}(\mathbf{o}_{b},\mathbf{o}_{t}) = |\mathbf{o}_{b} - \mathbf{o}_{t}|_{max}$. 
Also, we set $\lambda=0.9$. 

\begin{figure}[htb]
\centering
\includegraphics[width=1in]{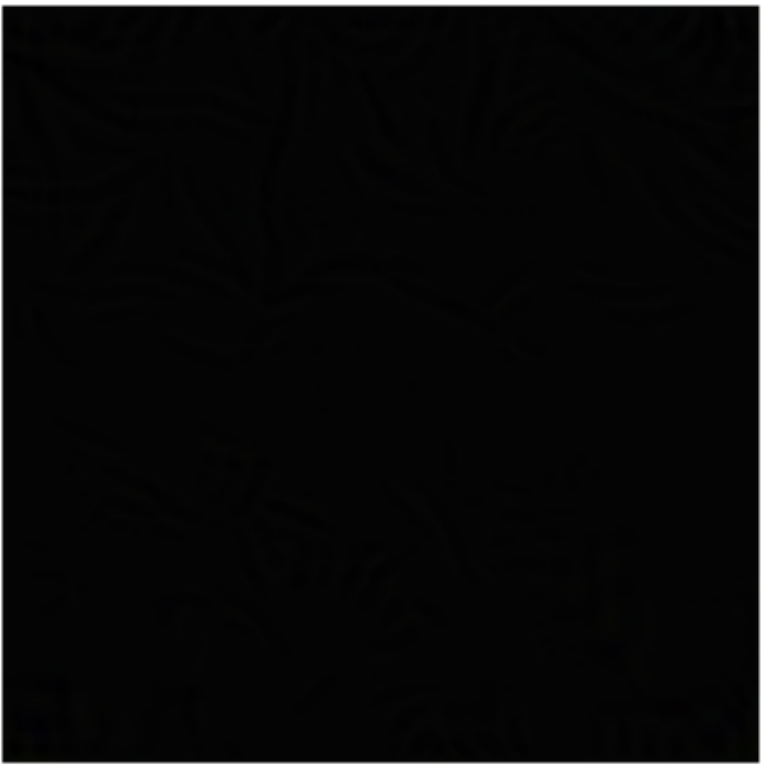}
\includegraphics[width=1in]{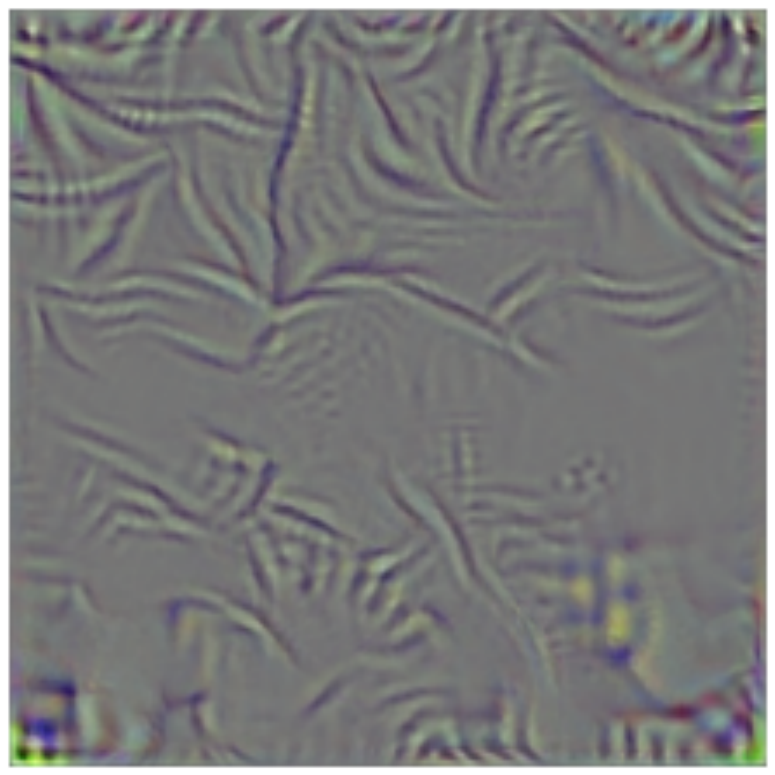}
\caption{Perturbed baseline before and after min-max normalization.}\label{f:baseline}
\end{figure}

After training we got a baseline that still looked black, although its pixel values actually were between
$-5.0$ and $5.25$ in a scale $0-255$, so the minimum pixel value $-5.0$ of the baseline was a bit outside the scale
minimum. After min-max normalization it looked like the right square in Figure~\ref{f:baseline}.

When the computed baseline was fed to the network the output vector was not yet perfectly
uniform but was much closer to being uniform than it was initially, with its components now
ranging from $0.0001$ to $0.0036$.  Some deviation from the mean is always expected and we
consider the improvement good enough for our purposes.

\begin{figure}[htb]
\centering
\includegraphics[width=2.72in]{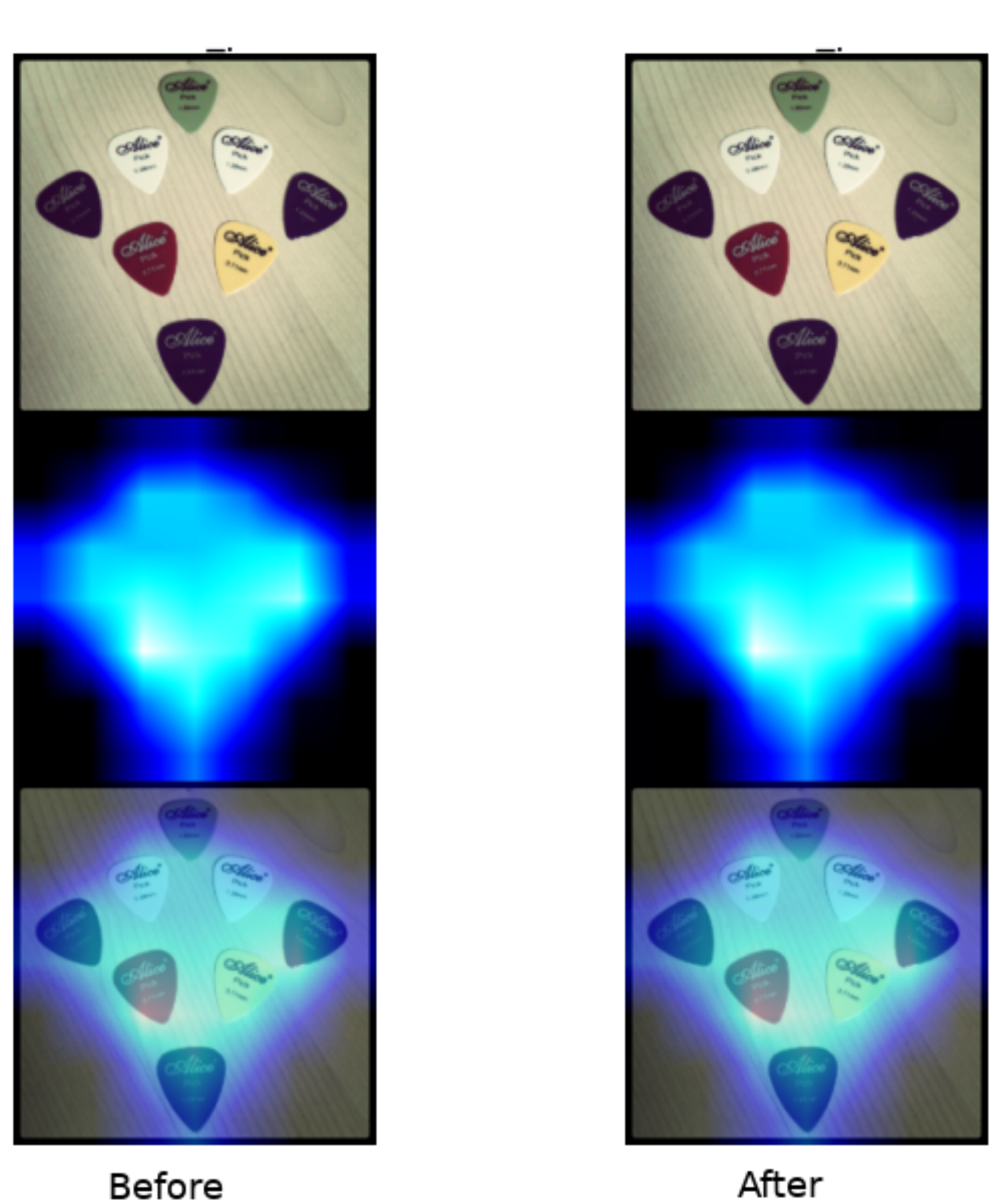}
\hskip 0.5in
\includegraphics[width=2.72in]{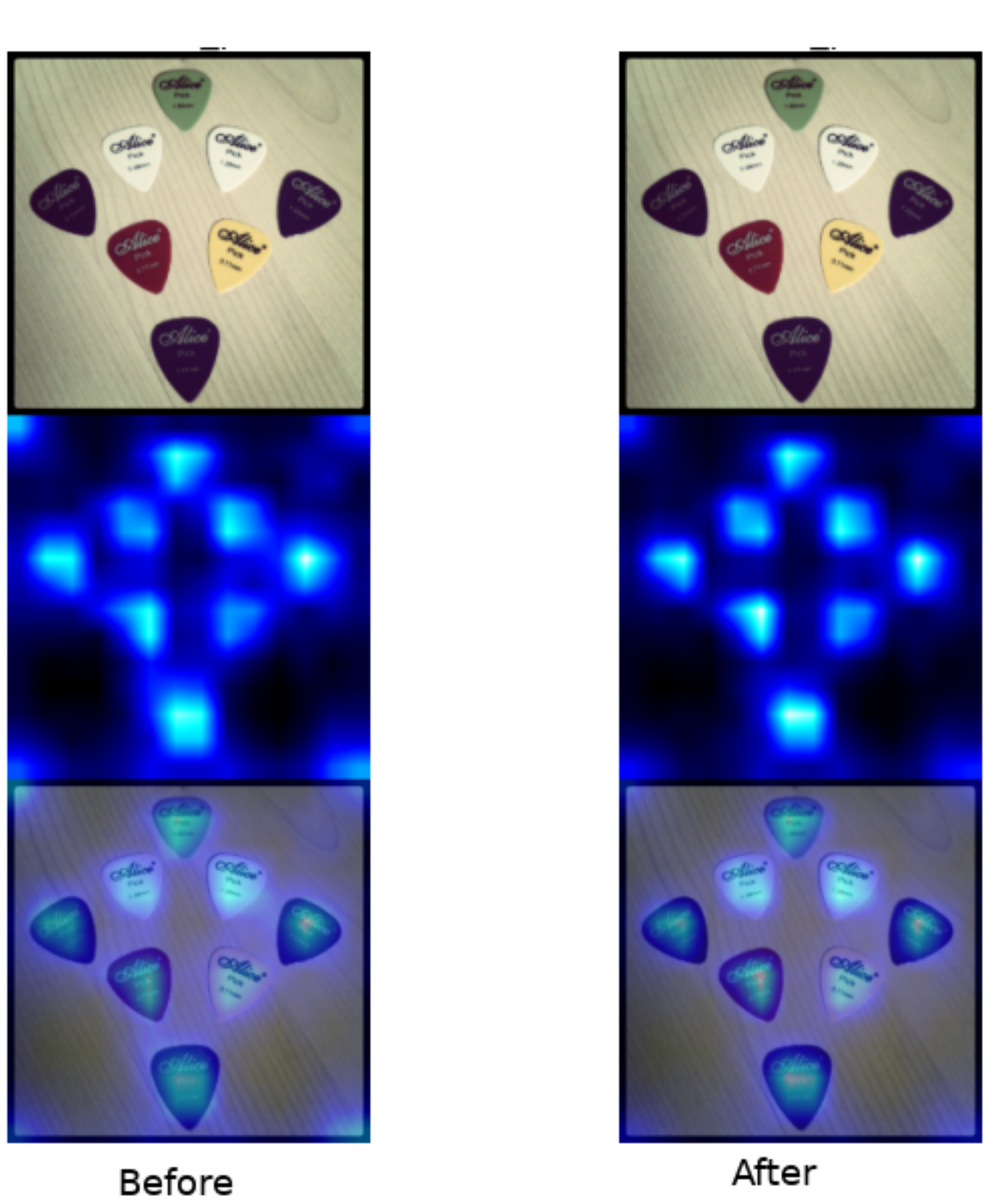}
\caption{Baseline before and after min-max normalization, produced at the last layer 
of the last convolutional block (left),
and the last layer of the next to the last convolutional block (right).}\label{f:heatmaps}
\end{figure}

Finally we show the heatmaps produced by our RSI-Grad-CAM
for an image from ImageNet (consisting of a set of picks)
before an after the baseline perturbation
(Figure~\ref{f:heatmaps}).
Each image contains the original image, the heatmaps produced, and an overlay of both.

As it can be seen, in this case the differences are minimal, although 
for the heatmaps produced at the next to the last block (right) 
the heatmap produced after the baseline perturbation looks slightly sharper
than before the perturbation.

\section{Conclusions}

We have described two gradient-based attribution methods that require to feed a network
from a baseline to a given input. The choice of baseline is to some extent arbitrary,
but we have shown that
after the initial choice it is possible to perturb it slightly so that when fed to
the network the output vector produced is an approximately uniform probability vector.
Finally, we have tested the perturbed baseline using our RSI-Grad-CAM attribution
method to produce heatmaps for an image of ImageNet using an VGG19 network. The outcome
is only slightly noticeable, but it serves as a proof of concept showing a promising
direction for future work.

\end{document}